\documentclass[conference]{IEEEtran}
\IEEEoverridecommandlockouts
\usepackage{cite}
\usepackage{amsmath,amssymb,amsfonts}
\usepackage{algorithmic}
\usepackage{graphicx}
\usepackage{textcomp}
\usepackage{xcolor}
\usepackage{subcaption}
\usepackage{float}
\def\BibTeX{{\rm B\kern-.05em{\sc i\kern-.025em b}\kern-.08em
    T\kern-.1667em\lower.7ex\hbox{E}\kern-.125emX}}

\usepackage{subfig}

\begin{document}

\title{Too Much of a Good Thing: \\ When sim2real Efforts Impede Policy Learning \\(And What to Do About It)
\thanks{
\textsuperscript{1} \text{Apptronik,} \textsuperscript{2} \text{University of Texas at Austin}
}
}

\author{
\IEEEauthorblockN{Kyle Morgenstein$\textsuperscript{1,2}$}
\and
\IEEEauthorblockN{Bharath Masetty$\textsuperscript{1}$}
\and
\IEEEauthorblockN{Stephen Welch$\textsuperscript{1}$}
\and
\IEEEauthorblockN{Luis Sentis$\textsuperscript{2}$}
}

\maketitle

\begin{figure*}[ht!]
\centering
\begin{subfigure}{0.3\textwidth}
  \centering
  \includegraphics[width=\linewidth]{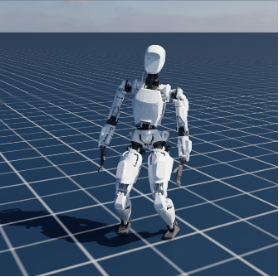}
  \caption{Policy Learning - IsaacLab}
  \label{fig:isaac}
\end{subfigure}\hspace{0.03\textwidth}%
\begin{subfigure}{0.3\textwidth}
  \centering
  \includegraphics[width=\linewidth]{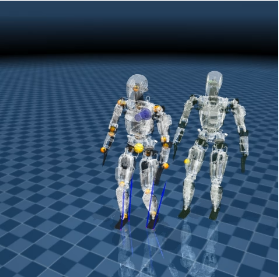}
  \caption{Sim-to-Real Transfer - MuJoCo}
  \label{fig:mj}
\end{subfigure}\hspace{0.03\textwidth}%
\begin{subfigure}{0.3\textwidth}
  \centering
  \includegraphics[width=\linewidth]{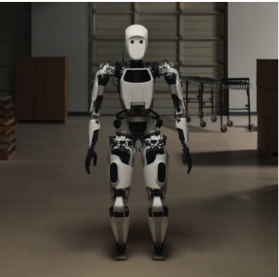}
  \caption{Hardware Deployment}
  \label{fig:hw}
\end{subfigure}
\caption{The sim2sim2real paradigm. A. Initial policy learning is performed in IsaacLab using a reduced order model. A forward model is learned concurrently. B. The policy is then transferred to MuJoCo via state-based imitation, using the forward model to generate references. C. The policy is then deployed on the Apptronik Apollo robot zero-shot.}
\label{fig:test}
\vspace{-1em}
\end{figure*}


\begin{abstract}
While sim2real efforts are necessary for effective policy transfer to hardware, there is such a thing as ``too much of a good thing''. We argue that sim2real efforts have led to misaligned incentives with policy learning, resulting in simulator lock-in and poor policy exploration due to the unreasonable constraints imposed by the real world. We offer a diagnosis and explanation of the current status of the problem, and propose a potential solution via a sim2sim2real paradigm that leverages the robot's kinematics as the sole design constraint.
\end{abstract}

\begin{IEEEkeywords}
sim2real, reinforcement learning, humanoids
\end{IEEEkeywords}

\section{Motivation}

Principled sim2real efforts require extensive experimentation and tuning to determine appropriate simulation parameters such as rotor inertia, joint friction, PD gains, control latency, and inertial properties.
Implementation differences between simulators and physics solvers mean that system identification results obtained for one simulator may not transfer identically to another, and system identification efforts may be tailored to the choice of simulator.
In practice, this results in simulator lock-in, as extensive engineering effort may be required to obtain system identification results for each desired simulator.
Moreover, while sim2real efforts focus on high-fidelity simulation of a robot's dynamics to facilitate zero-shot transfer, the resulting simulation model may impede effective policy learning \cite{hu2025impactstaticfrictionsim2real}.
We aim to decouple the relationship between behavior generation and system identification.
The remainder of this position paper are as follows:
First, we diagnose the ways in which the current sim2real paradigm impedes policy learning, both theoretically and as a matter organizational resource allocation.
Then, we detail a real-world case study in which both issues were overcome as part of the development cycle of the Apollo humanoid robot.
Finally, we make suggestions and predictions for effectively balancing sim2real concerns with policy learning efforts for large humanoid robots.

\section{The Current Policy Learning Paradigm is Overfit to sim2real}
The current dominant paradigm for deploying reinforcement learning controllers on hardware follows Fig. \ref{fig:sim2real_paradigm}.A.
After performing system identification, a policy is trained in simulation and then deployed on hardware zero-shot.
By comparing rollouts from simulation and the real robot, the robot model may then be refined, iteratively closing the sim2real gap.
This recipe has worked well in practice, enabling teams to push robots to their mechanical and electrical limits \cite{spotrun}.
However, this paradigm is overfit to sim2real concerns, and takes for granted that policy learning can find near-optimal behaviors with the high-fidelity system model.
We document at least two scenarios in which this assumption is overly restrictive.

\subsection{Simulator Lock-In Leads to Unnecessary Restrictions} 
Even small differences in system model can have large impacts on sim2real transfer.
In the past, teams overcame this rigidity via domain randomization such that the policy was robust over a family of system models with bounded variation.
However, it is now known that excessive domain randomization leads to conservatism in the policy \cite{adaptvrobust}, and so this strategy has fallen out of practice in favor of more rigorous system identification and state estimation efforts.
As a result, teams have opted to focus efforts on a single simulator, often termed a ``validation sim'' with known sim2real transfer dynamics.
This practice is not unique to learning, and has been used extensively with model-based controllers as well.
Thus, the choice of validation sim must be made carefully, as it must be able to support policy learning, validate the real kinematics and dynamics of hardware with sufficient fidelity for sim2real transfer, and oftentimes integrate with hardware-in-the-loop deployment infrastructure as well.
Once a validation sim is sufficiently mature, the engineering cost required to support or experiment with other simulators is often insurmountable.
This "simulator lock-in" can restrict the domain of possible research directions based on the abilities and limitations of the validation sim.

\subsection{Dynamic Feasibility is an Unreasonable Constraint} 
While training a policy in a high-fidelity simulator can enable zero-shot hardware deployment, enforcing dynamic feasibility as a strict constraint may hinder policy learning or otherwise render the task unsolvable.
In the current dominant policy learning paradigm for humanoid locomotion, desired PD targets are randomly sampled from a Gaussian distribution to facilitate policy exploration.
In lighter robots, this sampling procedure results in high frequency stuttering behavior early in training.
This stuttering quickly becomes high frequency foot steps, which enables the robot to quickly learn to stand and begin locomotion early in training.
For a larger, heavier humanoid, higher gains are required to track desired joint positions in the presence of gravity.
Following past works, let $K_p=I\omega_n^2$ be the motor stiffness \cite{raibertgain}.
The natural frequency is set by the length of the moment arm, $\omega_n \propto\sqrt{L^{-1}}$ and the reflected inertia is approximately proportional to the square of the moment arm, $I \propto mL^2 $, resulting in $K_p \propto mL$.
As a humanoid's size and weight grows, the gains required to compensate for gravity and track desired joint positions grows with both the mass and height of the robot.
However, assuming a fixed gear ratio, the maximum torque $\tau_\text{max}$ available from a standard brushless motor does not increase with size as quickly.
max If you naively scale a robot by a constant factor $c$, and note that mass scales with the cube of the scale factor, then we have $K_p \propto c^4$. However, $\tau_\text{max}$ scales cubically with the motor volume, so $\frac{K_p}{\tau_\text{max}} \propto c$, meaning as the size of the robot grows, the effective torque available normalized by the gains is reduced by $\frac{1}{c}$.
As a result, the exploration scheme used by RL techniques results in a core conflict: either set the gains appropriately for sim2real transfer, but suffer poor sample efficiency as the exploration scheme results in quickly saturating the torque limits, or intentionally set lower gains, allowing for better policy learning but with poor sim2real transfer.


\begin{figure}
    \centering
    \includegraphics[width=\linewidth]{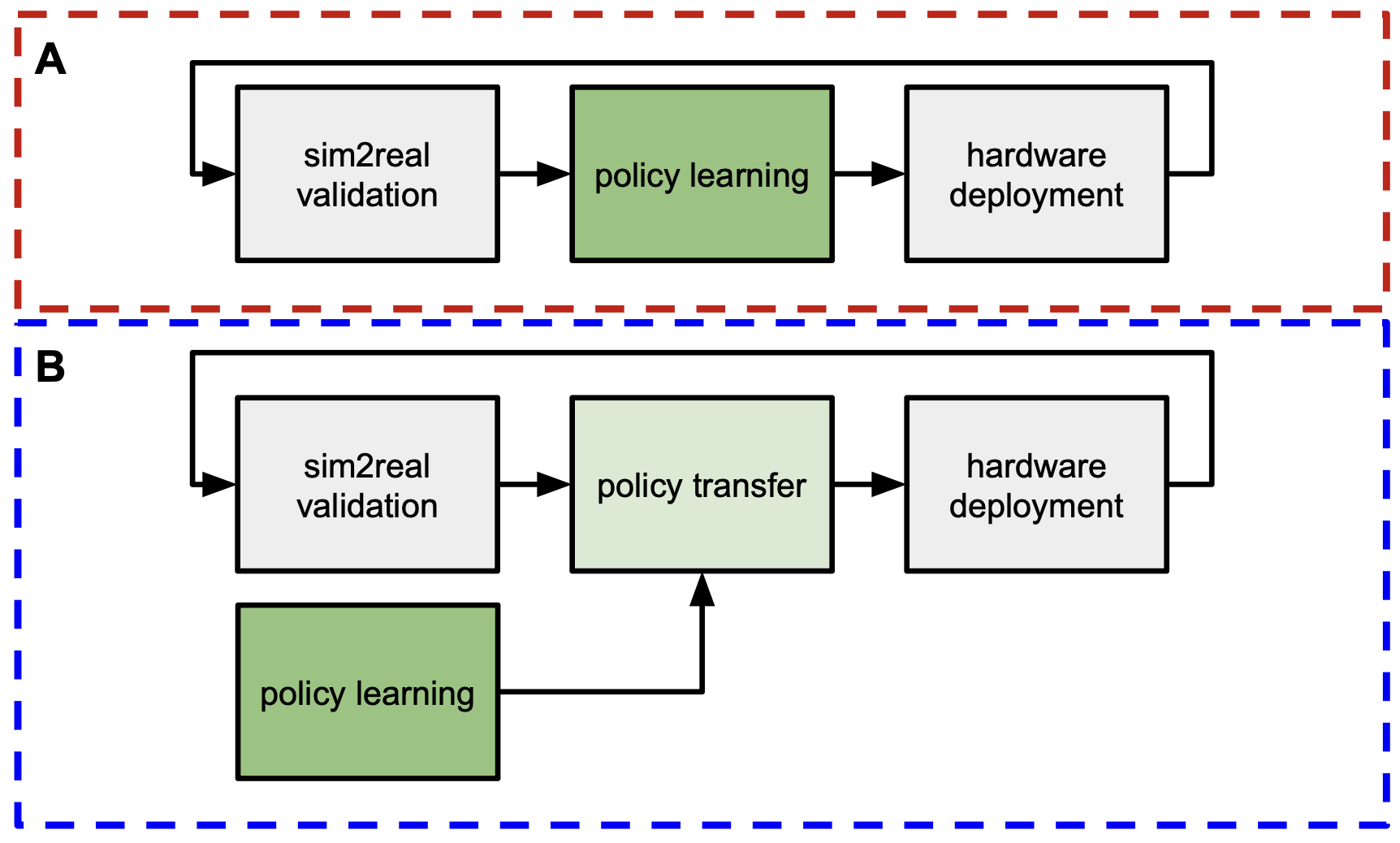}
    \caption{A. The current standard sim2real policy learning paradigm iteratively improves the simulation model of the robot and then performs policy learning on the updated model. The new policy is transferred to hardware zero-shot. B. The proposed sim2real policy learning and transfer paradigm separates policy learning for behavior generation from the core sim2real loop. The kinematically feasible motion is then transferred to the high-fidelity model via sim2sim kinematically guided policy transfer.}
    \label{fig:sim2real_paradigm}
    \vspace{-1em}
\end{figure}

\section{Kinematics Are All You Need (for Policy Transfer)}

\begin{figure*}[t]
    \centering
    \includegraphics[width=\linewidth]{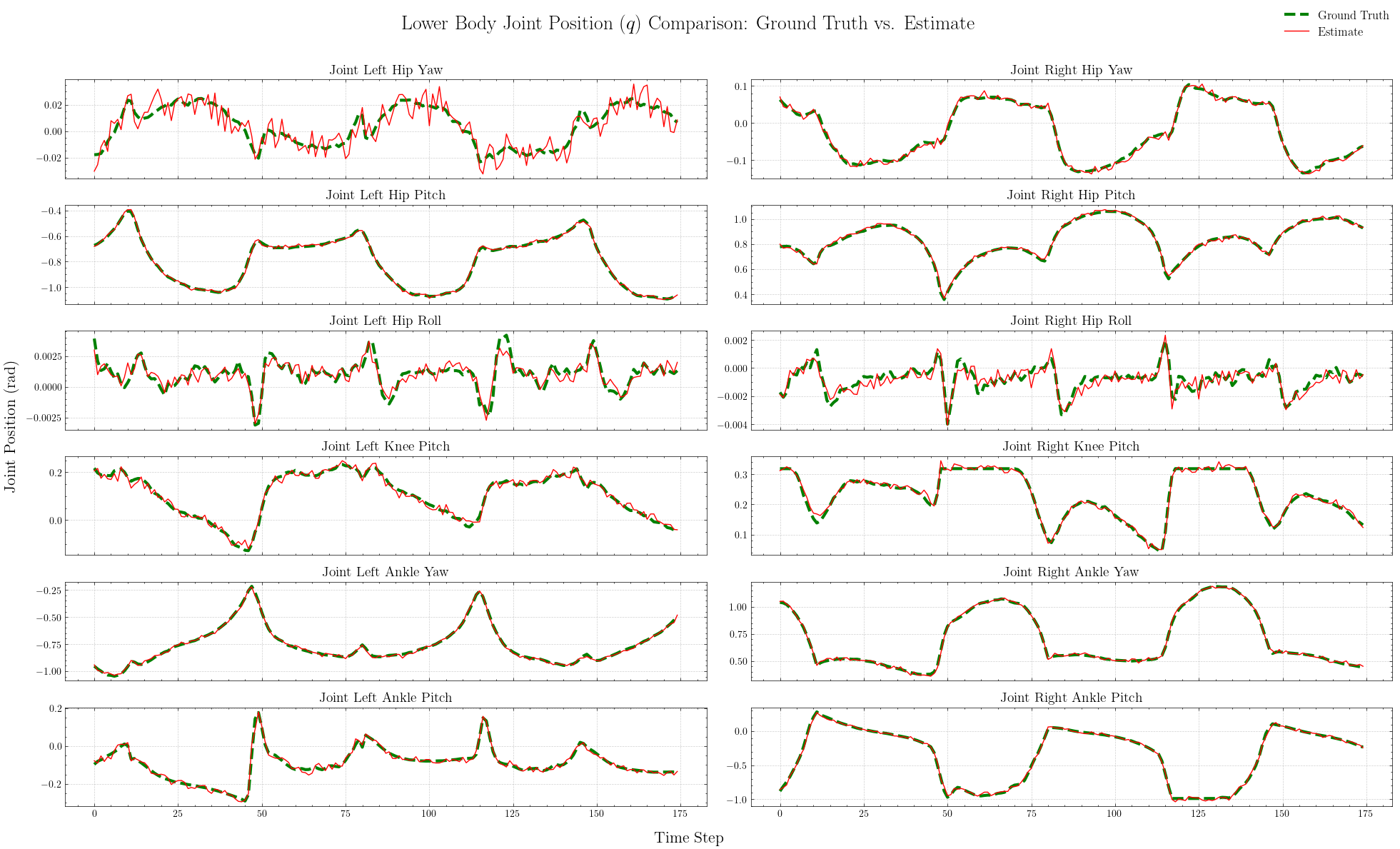}
    \caption{The forward model is trained to estimate the next kinematic state of the robot. The forward model observes the same state as the actor. The estimated lower body joint positions are plotted against the ground truth positions.}
    \label{fig:forward_model}
    \vspace{-1em}
\end{figure*}

Given the challenges with the current sim2real paradigm, we now propose an alternative recipe that 1) scales regardless of robot size 2) attempts to make policy learning easier and 3) vastly reduces the engineering overhead for testing new simulators and tools.
We start with the following formalization of the problem:

We would like to find a control policy for a robot with forward kinematics $FK(\cdot)$.
We define two simulation environments, a validation simulation that has been tuned to match the forward dynamics of the robot $FD(\cdot)$ as closely as possible, and a support simulation that represents an idealized environment for policy learning.
We assume that the validation simulation can also be used for policy learning.
The support sim may have the same physics engine as the validation sim (e.g. both based on MuJoCo or IsaacSim) but the dynamics of the support sim differ such that the action space of the policy differs.
For example, while the validation sim uses the system model found via system identification efforts, the model used in the support sim is more amenable for policy learning, e.g. using lower gains, no system latency, etc.
We formulate a policy learning paradigm that allows for easy transfer between the support and validation sims.

When transferring policies across environment, observation space, or simulator, student-teacher distillation \cite{studentteacher} and latent-guidance \cite{rma} methods have all proven effective.
However, prior methods fail to transfer policies across action spaces and robot dynamics due to the difference in required actuation strategies.
Similarly, reference motions may be embedded via RL with motion imitation strategies \cite{deepmimic, amp} in order to achieve the same behavior between different simulators, but these strategies require state or state-action trajectories a priori.
Instead, we propose a concurrent estimation and imitation framework that treats the robot kinematics as the shared ground truth between domains.
This choice allows us to train the teacher policy via methods that may not be suitable for sim2real transfer, whether due to fundamental or engineering constraints.
While the teacher policy is trained, a forward model is trained concurrently to estimate the next state of the robot.
Once the policy and forward model both converge, the forward model is used as a kinematic reference to transfer the teacher policy into the validation simulation that is more suitable for sim2real transfer.
A reinforcement learning policy can then be trained in the validation simulator using the trained forward model as an imitation signal to implicitly embed the teacher policy into the student via the desired kinematic state.
This technique vastly reduces the weight of selecting the ``correct'' validation simulator, as initial policies trained in support simulators can easily be transferred via the kinematic forward model guidance, and also allows for more effective policy learning, with exploration unconstrained by hardware limitations.
The proposed paradigm is depicted graphically in Fig. \ref{fig:sim2real_paradigm}.B.


\section{Case Study}
We train a locomotion policy in IsaacLab \cite{mittal_orbit_2023} using a reduced, low-fidelity model of the Apptronik Apollo humanoid robot, and then transfer the policy's behavior to an extensively validated full-order model of the robot in MuJoCo \cite{todorov_mujoco_2012} under different dynamics.

\subsection{Support: IsaacLab}
The IsaacLab model utilizes a reduced order representation of the robot that does not model the closed-loop kinematics of Apollo, allowing the action space of the environment to be an offset fed to a low-gain PD controller, which has been shown favorable for learning locomotion policies \cite{actionspace}. 
To further simplify the task, we choose not to model the joint friction in the idealized pin joint actuators, nor do we simulate system latency, joint torque, or velocity limits.
Thus, the IsaacLab setup can be understood to be a simplified, idealized setting for policy learning.
We use the standard observation space with commands given by base velocity, base height, and gait frequency.
We additionally observe the pose of each body link in the root frame.
The total observation is given by $o_t=\left[ v, \omega, \hat g, v_\text{cmd},h_\text{cmd},f_\text{cmd},q,\dot q,a_{t-1},r_\text{body} \right]$.
Both the policy and the forward estimator receive the same observation, while the critic receives additional privileged information that may not transfer well across simulators, such as ground reaction forces.
The forward model produces estimates for the next kinematic state and contact state $\hat o_{t+1} = \left[ \hat q, \hat{\dot q}, \hat r_\text{body}, \hat C_\text{foot}\right]$.
The policy is trained via PPO \cite{ppo} with an implementation based on RSL-RL \cite{rslrl} and the forward model is trained by minimizing the expected sum of the $L1$ and $L2$ losses to penalize both large and small magnitude estimator errors, as in \cite{l1l2, kyle}.
The estimation performance of the forward model can be seen in Fig. \ref{fig:forward_model}.

\subsection{Validation: MuJoCo}
The MuJoCo simulator models the full closed-loop kinematics of the robot and is validated against hardware.
The MuJoCo model converts reduced-order joint targets to full-order joint targets fed to a high-gain PD controller via forward kinematics.
The same gains are used in simulation and during hardware deployment.
The MuJoCo simulator was selected as the validation simulation due to its cross-platform compatibility, reduced computational requirements, ease of use, and ease of integration with existing tools and infrastructure.
However, in comparison to IsaacLab, MuJoCo currently lacks features for RL and integration with state-of-the-art learning libraries.
Extensive system identification efforts allow reliable zero-shot sim2real transfer for policies trained with the the MuJoCo model of the robot.
The proposed concurrent estimation \& imitation framework allows us to take advantage of the expressiveness and ease of experimentation afforded by IsaacLab without requiring repeated sim2real validation efforts nor abandoning the mature deployment infrastructure and tooling built around MuJoCo. To transfer the behaviors learned in IsaacLab to MuJoCo, we train the policy with state-based imitation rewards \cite{deepmimic} in a MuJoCo Playground-like environment \cite{mjplayground} via Brax \cite{brax} where the reference state is provided by evaluating the frozen forward model.
The observation space of the policy is similar to the teacher policy but excludes base linear velocity and includes sinusoidal phase variables. 
The critic may observe any additional relevant information such as feet air time, actuator output, etc.
After convergence, the policy can be deployed zero-shot on hardware.

\section{Conclusions}

We argue that the current paradigm of training policies with a single, high-fidelity robot model for zero-shot hardware deployment is overly restrictive, both from a policy learning perspective and an engineering perspective. By using a kinematic forward model as a shared representation for policy transfer between different robot dynamics models, we remove the restrictions a single model imposes and allow researchers and practitioners to use the dynamics model best suited to the task. This approach of synthesizing kinematic solutions and then refining them to be dynamically feasible is not new - in fact, it is standard in model-based control architectures.
Taking inspiration from such approaches, we tailor each simulation environment for either policy learning or sim2real transfer, simplifying both.
The conflict between mechanical dynamics, learning dynamics, and infrastructure is at the heart of what makes policy learning for robotics so difficult.
By balancing efforts to validate and robustify control approaches with efforts to adopt cutting edge techniques, organizations can more efficiently push the frontier of humanoid robotics.

\bibliographystyle{IEEEtran}
\bibliography{bib}
\vspace{12pt}

\end{document}